\documentclass[sigconf, nonacm]{acmart}

\AtBeginDocument{%
  \providecommand\BibTeX{{%
    \normalfont B\kern-0.5em{\scshape i\kern-0.25em b}\kern-0.8em\TeX}}}

\setcopyright{none}
\copyrightyear{}
\acmYear{}
\acmDOI{}

\newcommand*{\system}{SINGA-Easy}

\usepackage{algorithm}
\usepackage{algorithmicx}  
\usepackage{algpseudocode}
\usepackage{tikz}
\usepackage{color}
\usepackage{amsthm}
\usepackage{xcolor}
\usepackage{array}
\usepackage{amsmath,amsfonts}

\usepackage{graphicx}
\usepackage{subfigure}
\usepackage{textcomp}

\begin{document}

\title{\system{}: An Easy-to-Use Framework for MultiModal Analysis}


%\author{Paper ID:88}

\author{Naili Xing$^1$, Sai Ho Yeung$^1$, Chenghao Cai$^{1,4}$, Teck Khim Ng$^1$, Wei Wang$^1$, Kaiyuan Yang$^1$, Nan Yang$^1$, Meihui Zhang$^2$, Gang Chen$^3$, Beng Chin Ooi$^1$}


\affiliation{%
 \institution{$^1$National University of Singapore \country{Singapore} \hspace{8mm} $^2$Beijing Institute of Technology, China}
}

\affiliation{%
  \institution{$^3$Zhejiang University \country{China} \hspace{8mm} $^4$National University of Singapore (Suzhou) Research Institute \country{China}}
}

\email{{xingnl, yeungsh, caich, ngtk, wangwei, yangky, yangn, ooibc}@comp.nus.edu.sg}
\email{meihui\_zhang@bit.edu.cn}  \email{cg@zju.edu.cn}

\renewcommand{\shortauthors}{}

%%ww: I use --> xxx for replacing the word/phrase above with xxx
%%ww: I use ? for suggestions and ?? for questions.

\begin{abstract}
Deep learning has achieved great success in a wide spectrum of multimedia applications such as image classification, natural language processing and multimodal data analysis. Recent years have seen the development of many deep learning frameworks that provide a high-level programming interface for users to design models, conduct training and deploy inference. However, it remains challenging to build an efficient end-to-end multimedia application with most existing frameworks. Specifically, in terms of usability, it is demanding for non-experts to implement deep learning models, obtain the right settings for the entire machine learning pipeline, manage models and datasets, and exploit external data sources all together. Further, in terms of adaptability, elastic computation solutions are much needed as the actual serving workload fluctuates constantly, and scaling the hardware resources to handle the fluctuating workload is typically infeasible. To address these challenges, we introduce \system{}, a new deep learning framework that provides distributed hyper-parameter tuning at the training stage, dynamic computational cost control at the inference stage, and intuitive user interactions with multimedia contents facilitated by model explanation. Our experiments on the training and deployment of multi-modality data analysis applications show that the framework is both usable and adaptable to dynamic inference loads. We implement \system{} on top of Apache SINGA and demonstrate our system with the entire machine learning life cycle.

\end{abstract}




\maketitle

\section{Introduction}
\label{introduction}
Deep learning has been successfully adopted in a variety of multimedia applications such as image classification, speech recognition and news recommendation.
Driven by the increasing demand of real-world multimedia applications and an unprecedented growth of big data, many Deep Learning (DL) techniques and systems~\cite{shaofengdropout, wang2016database, wang2016effective} have been developed to facilitate the development of AI applications.
Although human-level performance has been achieved in areas like computer vision~\cite{DBLP:conf/icpp/YouZHDK18}, natural language processing~\cite{gardner2018allennlp} and speech processing~\cite{watanabe2018espnet}, the mass adoption of AI applications remains elusive due to two major challenges.

The first challenge is usability ~\cite{roh2019survey}. 
Many AutoML frameworks have been developed to improve usability.
These include Auto-WEKA \cite{DBLP:journals/jmlr/KotthoffTHHL17},
H20 AutoML \cite{H2OAutoML20}, Auto-Sklearn \cite{feurer2015efficient}, Auto-Pytorch \cite{DBLP:journals/corr/abs-2006-13799}, and Auto-Keras \cite{DBLP:conf/kdd/JinSH19}.
Among these frameworks, some provide functionalities for hyper-parameter tuning that can work with large datasets, and some support good user interaction and experience.
However, most of them do not take into account both.

Specifically, for hyper-parameters tuning, most DL model training processes focus on searching for the best hyper-parameter configuration using a set of stochastic gradient descent based optimization algorithms~\cite{bottou2012stochastic, bottou2010large, kingma2014adam}.
Many AutoML systems~\cite{liu2020admm, madrid2019towards} use Bayesian Optimization (BO)~\cite{pelikan1999boa, snoek2012practical} to tune the hyper-parameters automatically.
This is often time consuming as they need to evaluate many different combinations of hyper-parameters to obtain the best configuration.
To accelerate the searching process, AI-as-a-Service platforms such as Rafiki~\cite{rafiki} tune the hyper-parameters of the SGD algorithms in a distributed manner, but it pays little attention to the model (or architecture) related hyper-parameters.

In terms of user interaction and experience, DL systems typically hide implementation details and appear like a black-box to users.
To provide better interaction and user control, an easy-to-use APIs for managing the ML job in a finer-grained manner is required.
In addition, a good model explanation solution~\cite{selvaraju2017grad, ribeiro2016should, petsiuk2018rise} is also much needed in real-world applications, especially for those high-stakes applications.
For example, in the X-ray based medical diagnosis~\cite{ozcift2011classifier, foster2014machine, wang2020should}, making a wrong decision may lead to catastrophic consequences, and meanwhile, providing explainable AI solutions~\cite{holzinger2017we, goebel2018explainable, wang2019designing} can also engender user trust.
Auto-WEKA and H20 AutoML provide graphical user interfaces for datasets, models and task management.
In addition, H20 AutoML provides a number of model explanation functions based on variable importance and dependency.
However, most of these existing systems do not provide full support for automatic hyper-parameter tuning, good user interaction and model explanation.

The second challenge is adaptability -- inference services have to support elastic computation control in real-time, as it is not practical and often infeasible to scale the computational resources of the system to handling peak workloads.
In practice, the users may want to obtain more prediction results within a stipulated time and meanwhile can tolerate a slight decrease in accuracy.
A conventional static model takes a fixed amount of computation and thus can not trade off accuracy and efficiency dynamically, which therefore 
is unable to handle the fluctuating workload.
Clipper~\cite{Cliper2017} focuses on the prediction serving by introducing a new framework and explores several optimization techniques such as caching and model selection to improve latency and accuracy.
However, it simply drops the instances if they cannot be processed within the time limit in the presence of high workloads.
Model-Switching~\cite{ModelSwitching2020} - an online scheduler on top of Clipper, can select and switch to a different serving model based on the budget dynamically, which can achieve higher \emph{effective accuracy}.
Nevertheless, multiple models need to be trained beforehand and loaded to memory to support runtime model selection, which incurs additional overheads.
Notably, the ability to efficiently and effectively adapting the serving model size to the current workload and computational resources available is still lacking in most systems.

To address these two challenges, namely the usability and adaptability, an easy-to-use deep learning framework supporting automatic hyper-parameter selection, distributed training, dataset and model management, model explanation, and elastic computation control is required.
We design and implement such a system on top of Apache SINGA~\cite{ooi2015singa}.
The main contributions of this work are summarized as follows:

\begin{itemize}
% \vspace{-0.2cm}
\item We present an end-to-end open-sourced DL framework called \system{}, which is developed to facilitate the adoption of DL algorithms and inference services by domain-specific multimedia application users.
\system{} can automatically tune training jobs for pre-built complex models and adapt the model size when facing high inference workloads.
It also provides an intuitive APIs to manage the whole DL/ML life cycle and retrieve the inference result from various perspectives.

\item At the training stage, we propose a new auto-tuning framework combining both a distributed hyper-parameter tuning policy and an adaptive regularization method, which reduce the effort required for training an efficient and accurate model.
We also integrate to our system a novel technique to adapt the serving model size dynamically.

\item At the inference stage, we focus on the evaluation metric \emph{effective accuracy} and propose a new scheduling algorithm to adapt the model size to the current workload in real-time for achieving higher \emph{effective accuracy}, and meanwhile satisfy user-defined response time requirement under the available computational resources.
The algorithm also reduces manual effort in the deployment, scaling and workload balancing of the service.

%%% ooibc:
%%% every sentence must be grammatically correct -- it follows the structure subject, verb, object
\item To facilitate the adoption of multimedia applications, we integrate commonly used algorithms and provide an easy-to-use APIs. We also provide an option for users to evaluate model performance from the model explanation perspective provided by LIME~\cite{ribeiro2016should} and Grad-CAM~\cite{selvaraju2017grad}. 
%%Section\ref{sec:inference}

\item We demonstrate the usability and adaptability of our \system{}
%%over baselines
by conducting experiments on various real-world datasets and showcasing several multimedia applications.

% \item We demonstrate the usability and adaptability of \system{}
% %%over baselines
% by conducting experiments on various multimedia datasets.
\end{itemize}

The remainder of the paper is structured as follows.
Section~\ref{sec:architecture} introduces the system architecture and the dataflow between the system components.
Section~\ref{sec:inference} introduces the dynamic model serving framework.
%%and the model slicing and the scheduling algorithm in the \emph{Predictor} and \emph{Inference Worker}.}
%%% ooibc: need to get the simple grammar right -- singular/plural etc, present/past tense
Section~\ref{sec:userInteraction} presents system usability.
%with APIs examples and part of integrated modes of \system{}.
%% ooibc
Section~\ref{sec:experiment} presents the experimental study.
%to evaluate the effectiveness of \system{}.
We review the related work in Section~\ref{sec:related} and conclude the paper in Section~\ref{sec:conclusions}.

\begin{figure}
\centering
\includegraphics[width=8cm]{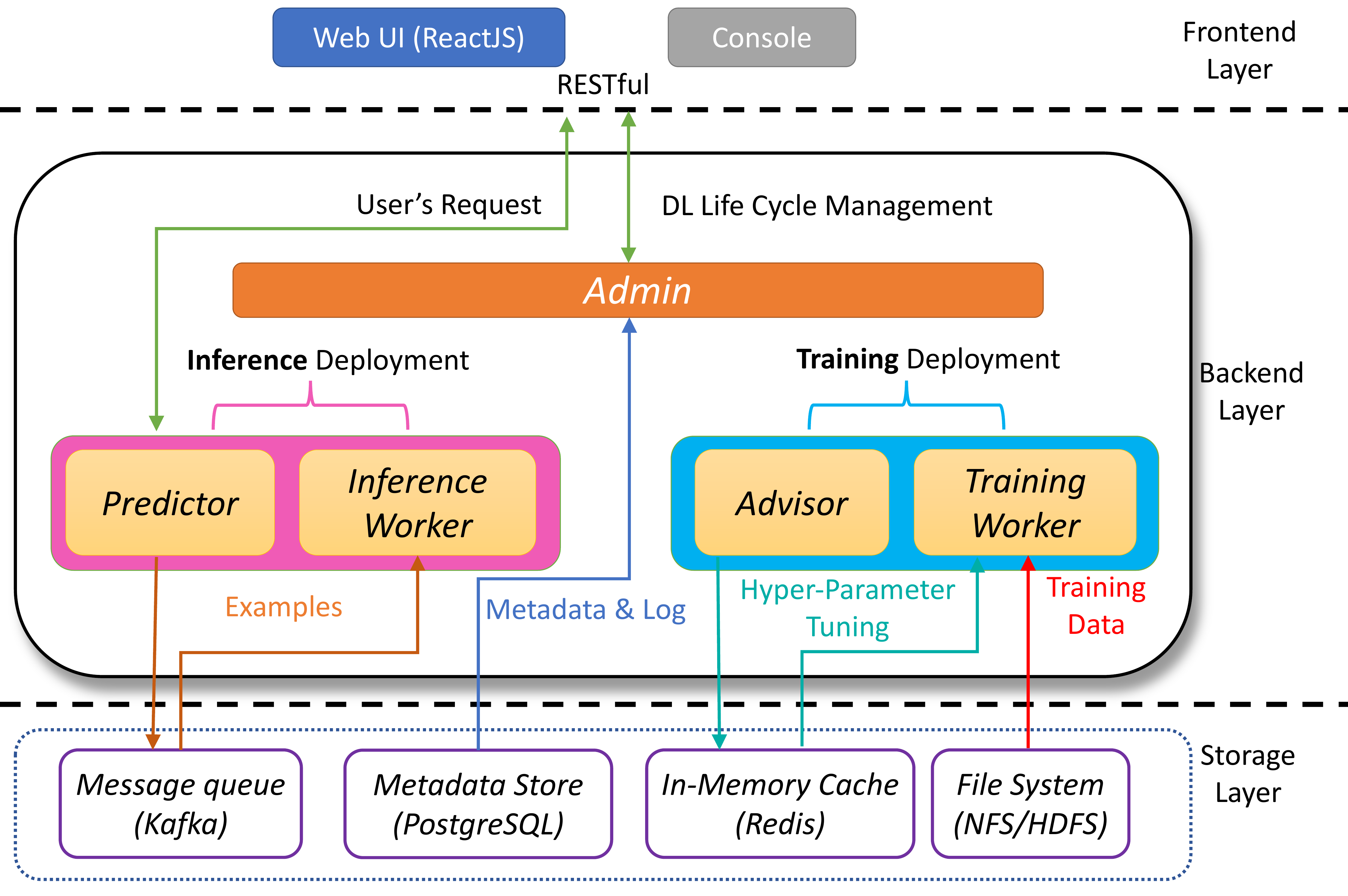}
\caption{\system{} system architecture overview.}
\label{fig:architecture}
\end{figure}

\section{System Architecture}
\label{sec:architecture}

This section introduces the system architecture of \system{}.
The software stack is illustrated in  Figure~\ref{fig:architecture} and  Figure~\ref{fig:software_stack} respectively.
The system consists of the
\emph{frontend layer}, the \emph{backend layer}, and the \emph{storage layer}.
Specifically, the frontend layer provides different HTTP APIs to manage both data and tasks.
Users can interact with the framework via either \emph{Python SDK-Client} or \emph{Web UI}.
In the following subsections, we will introduce the other layers in detail.

\subsection{Backend Layer}

\system{} is built on top of the base architecture of Apache SINGA.
The backend of the overall system comprises five essential components: \emph{Admin}, \emph{Training Worker}, \emph{Advisor}, \emph{Predictor} and \emph{Inference Worker}.

\emph{Admin} is the core component of the system's control plane, which exposes HTTP APIs for users to manage the whole ML lifecycle.
Upon receiving requests from users via RESTful APIs, it deploys a number of workers for model training and serving, and stores information of the user-defined tasks into a \emph{Metadata Store}. 
When a worker is deployed, the worker pulls the information of the task from the \emph{Metadata Store} and starts the task.

\begin{figure}
\centering
\includegraphics[width=8cm]{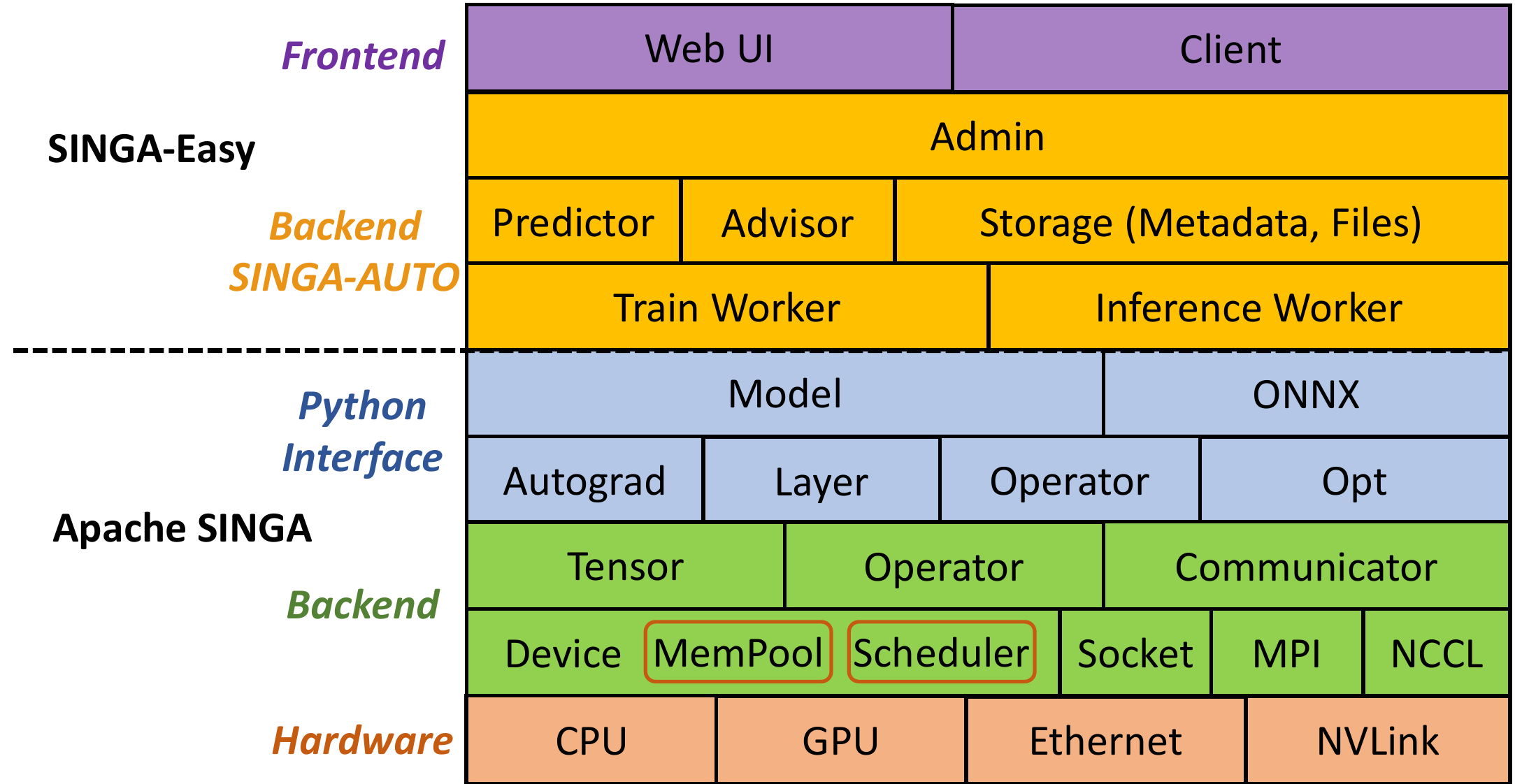}
\caption{\system{} software stack.}
\label{fig:software_stack}
\end{figure}

\emph{Training Worker} trains models by conducting trials proposed by a corresponding \emph{Advisor Worker}.
The computational kernel of \emph{Training Worker} supports various DL libraries in addition to Apache SINGA e.g.,  PyTorch~\cite{paszke2019pytorch}, and TensorFlow~\cite{abadi2016tensorflow}.
Figure ~\ref{fig:software_stack} shows the stack diagram with Apache SINGA as the DL framework,
%%%ww: since SINGA-Easy is a framework. You can use "engine" or "library" for SINGA, TF...
where the upper layers are constructed based on the lower layers.
For example, the component \emph{model} is defined using \emph{Layer}, \emph{Autograd}, \emph{Opt} and \emph{Operator}, etc.
Each of them is built on top of the basic data structures
%%with basic infrastructure
%%%ww: built on top of the basic data structures
of Apache SINGA such as \emph{Tensor} and \emph{Communicator}. 
We introduce the technical details of the training part in Section~\ref{sec:training}.
%%%ww: try to avoid using "will" when you can use the simple present tense. 
%%n l：OK

\emph{Advisor} performs hyper-parameter tuning by conducting multiple trials on a training job.
In each trial, it proposes the training configuration, i.e., knobs of the model and training algorithm, to be used by \emph{Training Worker}.
For the implementation of the Advisor, we adopt the Bayesian Optimization of Scikit-Optimize toolbox\footnote{Scikit-Optimize: https://scikit-optimize.github.io/stable}.

\emph{Predictor} is designed for ensemble modelling, which stands between users and \emph{Inference Workers}.
\emph{Predictor} receives requests, e.g., one or many images to be classified, from the users, then forwards the requests to a number of \emph{Inference Workers} and collects the prediction results.
\emph{Inference Worker} manages trained models for the inference jobs, which receives the request forwarded by the \emph{Predictor} and performs prediction.
The technical details of inference are discussed in Section~\ref{sec:inference}.

\subsection{Storage Layer}

The storage layer contains the following components for caching and storage of data:

\emph{Metadata Store} is a centralized and persistent database used to store the metadata of the whole system such as user metadata, job metadata, worker metadata and model templates. We adopt PostgreSQL\footnote{PostgreSQL: https://www.postgresql.org/}  in our system.

\emph{In-Memory Cache} is an in-memory data structure store used for fast asynchronous communication between \emph{Training Workers} and \emph{Advisor} at the training stage. Redis\footnote{Redis: https://redis.io} is used as the in-memory data store.

\emph{Message Queue} is a file-based data store used for supporting asynchronous communication
%%%ww: what is "asynchronizes"?
between \emph{Inference Workers} and \emph{Predictors} for Inference Jobs.
Apache Kafka\footnote{kafka: https://kafka.apache.org} is used in our system.

% \vspace{-0.3cm}
\subsection{Workflow}
\label{sec:workflow}

\system{} allows users to manage the whole ML life cycle and retrieve the prediction results from inference services.
The user firstly uploads the model, dataset, or annotation file to \emph{Admin}, which will be stored into a distributed storage (NFS). The metadata will be stored into \emph{Metadata Store}.
When use send requests to start training, \emph{Admin} launches one \emph{Advisor} and multiple \emph{Training Workers}.
\emph{Advisor} stores training configurations into \emph{In-Memory Cache}, and \emph{Training Workers} will conduct training accordingly.

In each iteration, \emph{Training Workers} 
report the training accuracy to \emph{In-Memory Cache}, which is then used by \emph{Advisor} to generate new configurations for the next training iteration.
After completing a training job, one \emph{Predictor} and multiple \emph{Inference Workers} will be created. The user can retrieve the \emph{Predictor} inference service's URL from \emph{Admin} to use the model inference services.
All these components communicate with each other via a message queue.

\subsection{Elastic Inference}
% sf: check

% \subsection{Distributed Training}
\label{sec:training}

\begin{figure}
\centering
\includegraphics[width=7cm]{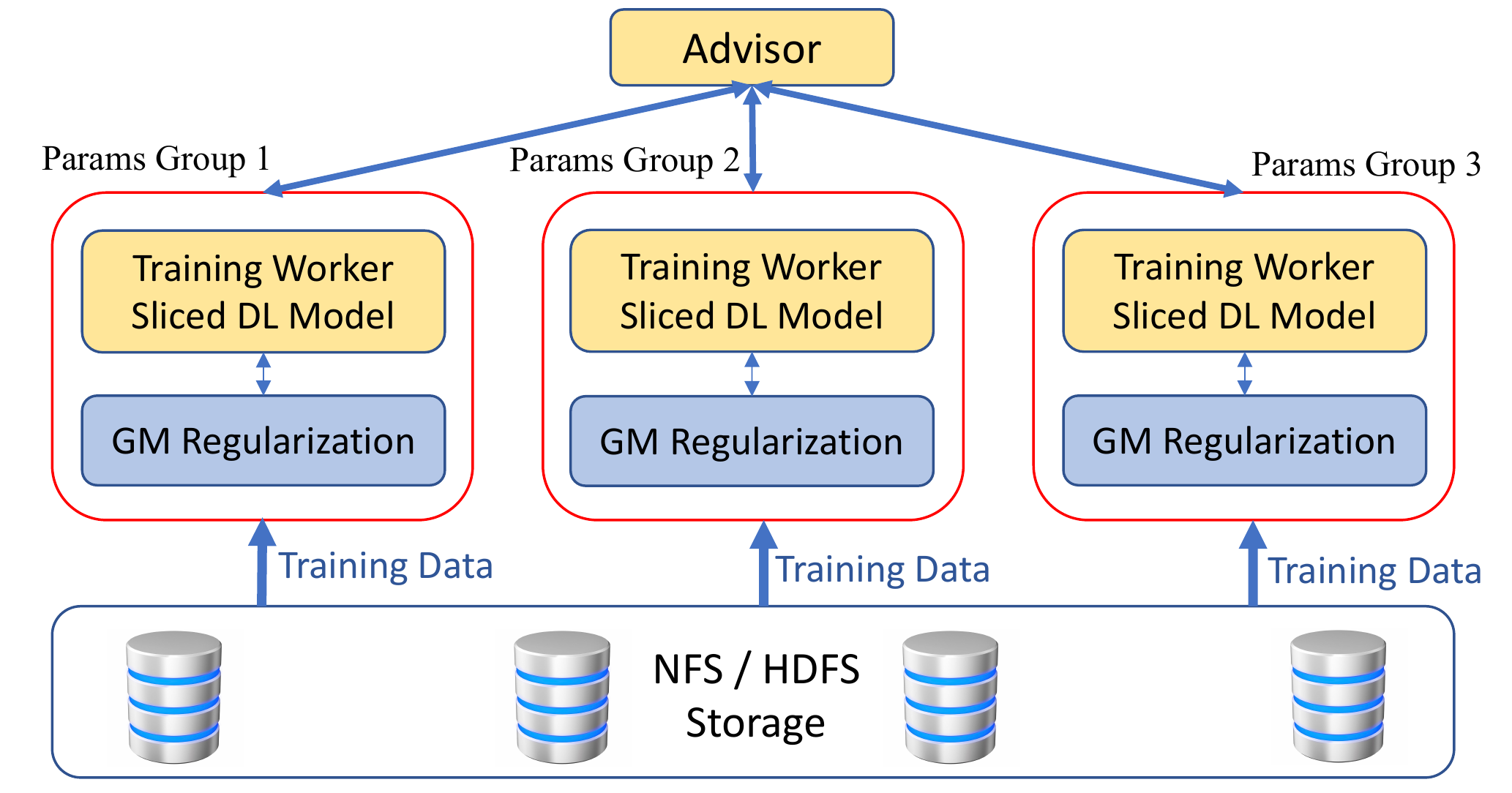}
\caption{Training Worker: distributed model training.}
\label{fig:training}
\end{figure}

Model slicing~\cite{cai2019model} is a general technique to enable deep learning models to support elastic computation.
Specifically, each layer of the model is divided into equal-sized contiguous computational groups.
During both training and inference, there is a single parameter slice rate $r$ that dynamically controls the fraction of groups involved in the computation for all the layers in the model, namely the model width.
In particular, these groups are trained dynamically to build up representations residually. The first group learns the base representation. Each subsequent group learns on top of all its preceding groups.
As a result, during inference, we can support accuracy-efficiency trade-offs by dynamically slicing a subnet of a certain width, where only the parameters of the activated groups are involved in computation.
Theoretically, the number of parameters and computation measured in FLOPs are both roughly quadratic to the slice rate $r$~\cite{cai2019model}, e.g., a slice rate of 0.5 can achieve up to four times speedup.
Therefore, we can support elastic inference by introducing the model slicing technique to the training stage.
Specifically, we can train the model with multiple slice rates beforehand. At the inference stage, the model can be switched to different sub-models adaptively.

The overall training process is illustrated in Figure~\ref{fig:training}.
We train the model with model slicing to render the ability of elastic computation.
For efficiency, the system trains the sub-models in a distributed manner by training these sub-model instances of different slice rates in a pool of workers.
After a few training iterations, all workers merge their local copy of weights and update them globally.
~\system{} reuses the distributed hyper-parameter tuning component of Apache SINGA and adds an adaptive Gaussian Mixture (GM) regularization technique~\cite{luo2018adaptive} to further improve the prediction performance of the model.

\section{Dynamic Model Serving}
\label{sec:inference}

\begin{figure}
% \setlength{\abovecaptionskip}{-0.1cm}
% \setlength{\belowcaptionskip}{-0.5cm}
% \vspace{-0.2cm}
\centering
    \subfigure[Single model serving]{
        \begin{minipage}[a]{0.5\textwidth}
        \includegraphics[width=8cm]{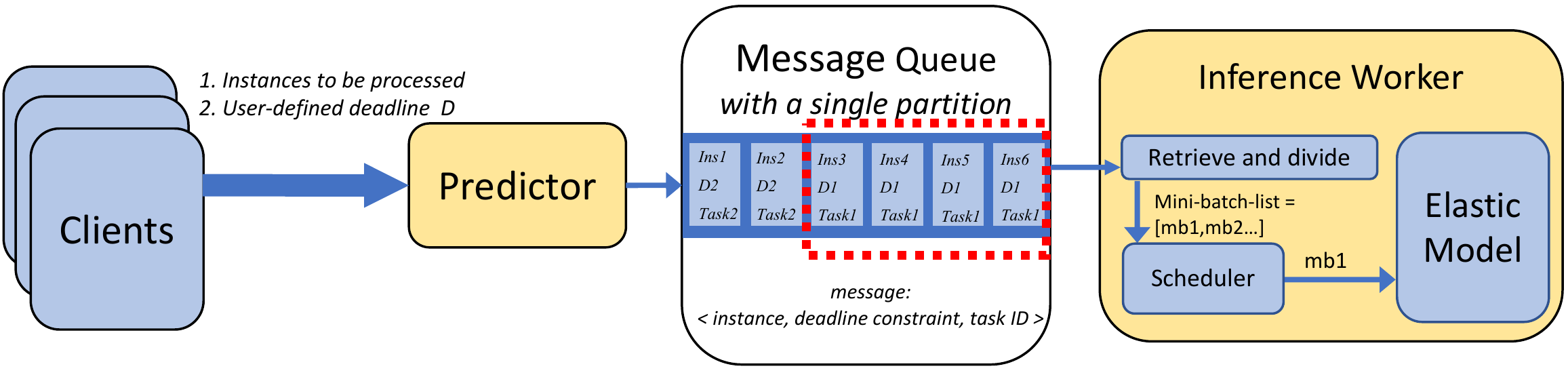}
    \end{minipage}}
    \subfigure[Multiple model serving]{
        \begin{minipage}[b]{0.5\textwidth}
        \includegraphics[width=8cm]{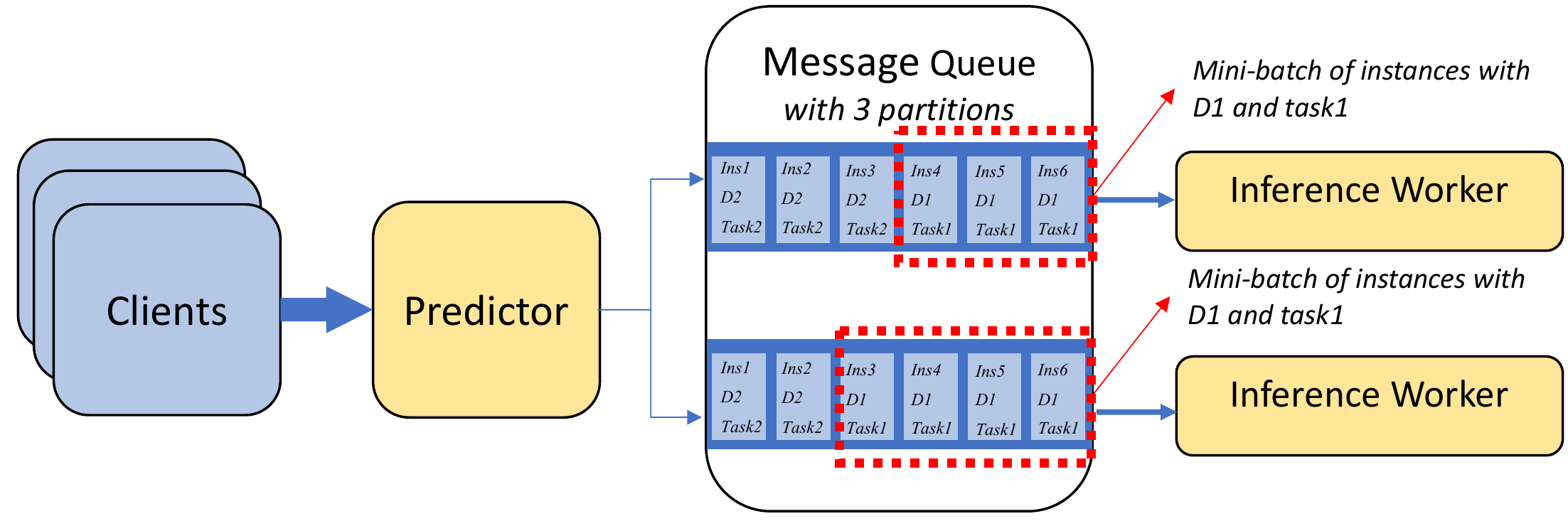}
    \end{minipage}}
% \caption{Inference stage: Each instance in the message queue contains the value and its deadline constraints.} 
\caption{Inference stage: each message in the message queue contains one instance and the global deadline constraint for this prediction task.} 
% sf: value means what?
% NL: value is the value of instances, like matrix for images, etc
\label{fig:inference2scen}
\end{figure}

To support dynamic model serving, we further propose a scheduling algorithm based on the elastic inference enabled via model slicing.
At the inference stage, the user can send multiple requests, each of which corresponds to one prediction task.
A prediction task may contains multiple instances to be processed by the serving model.
As shown in Figure \ref{fig:inference2scen}(a), each \emph{Client} sends instances and a global deadline constraint $ D $ to \emph{Predictor}.
The \emph{Predictor} and \emph{Inference Workers} will then work cooperatively to produce the inference result.
The \emph{Predictor} wrap each instance to be a message in the form of 3-tuple $<Instance, D, TaskID>$, it then pushes all messages to a message queue as a producer.
The \emph{Inference Workers} read message of the same prediction task from the queue as consumers, divide the instances into mini-batches, adjust the model size according to the scheduling algorithm and then conduct the inference.
The mini-batch is the minimum scheduling unit, which can contain one or more instances, e.g., one video for the video objection detection task or several texts for the text classification task.
To fully utilize the GPU parallel computing capability, the size of the mini-batch is typically set to the largest possible.
The algorithm is outlined in Algorithm~\ref{algo:ms}.
For ease of reference, we summarize all the variables used in subsequent sections in Table ~\ref{tab:variables}.

% sf: notations
% f^T not in table 1

To evaluate the effectiveness of the proposed dynamic model serving, we adopt the \emph{effective accuracy}~\cite{ModelSwitching2020} as the evaluation metric, which is fraction of correctly processed instances returned before a predefined deadline.
Specifically, denoting the prediction accuracy of a given model as $p$ and the fraction of instances that can be processed by the model before the deadline $D$ as $f^T$, \emph{effective accuracy} $p^{eff}$ can then be computed by $ p^{eff}=p * f^T $.
Briefly, \emph{effective accuracy} takes into account both the accuracy and efficiency of the serving model, as $p$ represents the prediction accuracy. Given the deadline, $f^T$ is determined by the efficiency of the model inference.
Notably, $f^T$ can also be seen as the throughput within the given time frame.
The goal of the scheduling algorithm is formally defined as follows.

\emph{Given the number of instances to be processed $N$ and a global deadline constraint $D$, the scheduling algorithm is to maximize the effective accuracy under the available computational resources in the model serving system}.

% To better evaluate the scheduling algorithm from both precision and throughput aspects, we use the \emph{effective accuracy} measure defined in Model-Switching~\cite{ModelSwitching2020}. A given model can be pre-characterized with the accuracy $p$ and the fraction $f^T$ of requested instances that meet the deadline constraints $D$. $f^T$ can also be seen as the throughput within a given time frame. The \emph{effective accuracy} is computed by $ p^{eff}=p * f^T $, and the goal of the algorithm is defined as follows.

% \emph{Under limited computational resources, given the number of instances $N$ and global deadline constraint $D$, the scheduling algorithm will maximize the overall \emph{effective accuracy} to satisfy $D$}.

% When the desired throughput and accuracy cannot be achieved simultaneously, the scheduling algorithm will attempt to prioritize throughput over precision.

% When the desired throughput and accuracy cannot be achieved at the same time, the scheduling algorithm will attempt to prioritize throughput over precision.
We develop two scheduling algorithms in two respective scenarios,
where the available computational resources can support either only one model in Section~\ref{sec:singlemodelinference} or multiple models in Section~\ref{sec:multimodelinference}.
In what follows, we denote that sub-model $m_i$ indexed by the slice rate $r_i$ has a prediction accuracy $p_i$ and takes $t_i$ time on average to process a mini-batch of a fixed number of instances.
% When the desired throughput and accuracy cannot be achieved at the same time, the scheduling algorithm will attempt to prioritize throughput over precision.
% We illustrate the scheduling algorithm using two scenarios
% where computational resources are limited to satisfy either only a single model in Section~\ref{sec:singlemodelinference} or multiple models in Section~\ref{sec:multimodelinference}.
% In what follows, we assume that any sub-model $m_i$ selected with slice rate $r_i$ has fixed precision $p_i$ and fixed processing time $t_i$ on each mini-batch. The precision is evaluated on validation data. 
%%%ww: need to add a note on this assumption. e.g., In practice, for each sub-model, we use the precision on the validation data as the initial precision, and update it  periodically using online data (if labels are available).
Typically, the full model (i.e., $r_i = 1.0$) has the highest accuracy while the lowest efficiency.
For both scenarios, there are $K$ sub-models trained beforehand.
The $N$ instances are divided into $N_{mb}$ mini-batches (each with $ \frac{N}{N_{mb}} $ instances) by \emph{Inference Worker}. The sub-model $m_i$ takes $T_i$ to process all $N_{mb}$ mini-batches.
% We assume that $N$ instances are divided into $N_{mb}$ mini-batches by \emph{Inference Worker} and sub-model $m_i$ requires $T_i$ to process $N_{mb}$ mini-batches.
% In general the full model where $r_i = 1$ has the highest accuracy but it has the slowest speed.
% We further assume that $K$ sub-models are trained. The \emph{client} sends $N$ instances to retrieve prediction results and also determines the expected deadline constraint $D$. We assume that $N$ instances are divided into $N_{mb}$ mini-batches by \emph{Inference Worker} and sub-model $m_i$ requires $T_i$ to process $N_{mb}$ mini-batches.
%%%ww: be careful about "mini-batches" vs "instances". Which one do you use in the scheduling algorithm?
%% NL: it's mini-batches, instances is sent by user, and we divide them to many mini-batches internally.
% The scheduling algorithm will generate the best scheduling policy denoted by $[n_1...n_m]$ where $n_i$ means there are $n_i$ mini-batches assigned to sub-model $m_i$.
Formally, the scheduling algorithms are to determine the best scheduling policy denoted as $[n_1...n_m]$, where $n_i$ is the number of mini-batches assigned to sub-model $m_i$.

\begin{table}
  \caption{Summary of variables.}
  \label{tab:variables}
  % \vspace{-0.5cm}
  \begin{tabular}{ll}
  \toprule
  Symbol & Meaning \\
  \midrule
  $m_i$ & The $i$-th sub-model  \\
  $r_i$ & Slice rate of $m_i$ \\
  $p_i$ & Accuracy of $m_i$ \\
  $t_i$ & Time for $m_i$ to process a mini-batch \\
  $p_i^{eff}$ & \emph{Effective accuracy} of $m_i$\\
  $K$ & Number of trained sub-models\\
  $n_i$ & Number of mini-batches assigned to $m_i$ \\
  $\widehat{W_i}$ & The maximum workload$^*$ that $m_i$ can handle\\
  $N$ & Number of instances in one prediction task\\
  $S_{mb}$ & Number of instances in a mini-batch, \\
  & which is fixed on each prediction task.\\
  $N_{mb}$ & Number of mini-batch, $N_{mb}=N~/~S_{mb}$ \\
  $D$ & User-defined deadline to process $N$ instances \\
  $W_{exp}$ & Expected workload of the system, $W_{exp}=N~/~D$ \\
  $T_i$ & Time spent by $m_i$ to process $N$ instances \\
  $T_{fast}$ & $T_i$ of the fastest sub-model \\
  $T_{slow}$ & $T_i$ of the slowest sub-model \\
  \bottomrule
\end{tabular}
\leftline{*Workload: the number of instances to be processed per second.}
%%%ww: how is workload defined? num of examples or num of mini-batches?
  %% nl: since user can only send the examples. here we use num of examples, we divide the exaplesinto mini-batches, so it's the same as num of mini-batches, 
% sf: a serious error here: $\widehat{W_i}=1~/~t_i * S_{mb}$
% \vspace{-0.8cm}
\end{table}

\renewcommand{\algorithmicrequire}{\textbf{Input:}}
\renewcommand{\algorithmicensure}{\textbf{Output:}}

\begin{algorithm}
    \caption{Model Serving \emph{Predictor} and \emph{Inference Worker}}
    \label{algo:ms}
    \begin{algorithmic}[1] 
        \Require $\mu = [m_1,\ldots,m_K]$, $\rho = [p_1,\ldots,p_K]$, $S_{mb}$, user's requests
        \Ensure Prediction results
        \Function {Predictor}{}
            \While{True}
	            \State Receive user's request with instances, $ D $
	            \State generate task ID for this prediction task
	            \State message: $<Instance, D, TaskID>$ $\gets$ each instance
	            \State Send the messages to the message queue
	        \EndWhile
        \EndFunction
        \Function{Inference Worker}{$\mu, \rho, S_{mb}$}
            \State Get $\tau = [t_1,\ldots,t_K]$ according to $S_{mb}$
	        \While{True}
	            \State Retrieve $ N $ messages with the same ID from the queue
	            \State Retrieve instances and $ D $ from message tuples
                \State Divide the instances into $ N_{mb} $ mini-batches of size $ S_{mb} $
                \State Store the mini-batches as a list $ \gamma $ 
	            \State $\eta, p^{eff}(\eta) \gets \texttt{Scheduler}(\rho, \tau, D, N_{mb}) $ 
	            \For{$n_i$ in $\eta$}
	                \For{j in $[1,\ldots,n_i$]}
	                    \State $ R \gets \mu[i].\texttt{prediction}(\gamma.pop()) $
	                    \State Send $ R $ as prediction results to the user
	                \EndFor
	            \EndFor
            \EndWhile
        \EndFunction
    \end{algorithmic}
% \vspace{-0.2cm}    
\end{algorithm}

% \vspace{-0.4cm}
\subsection{Single Model Serving}
\label{sec:singlemodelinference}
\begin{figure}[t!]
% \setlength{\abovecaptionskip}{-0.05cm}
% \setlength{\belowcaptionskip}{-0.6cm}
% \vspace{-0.4cm}
\centering
\includegraphics[width=8cm]{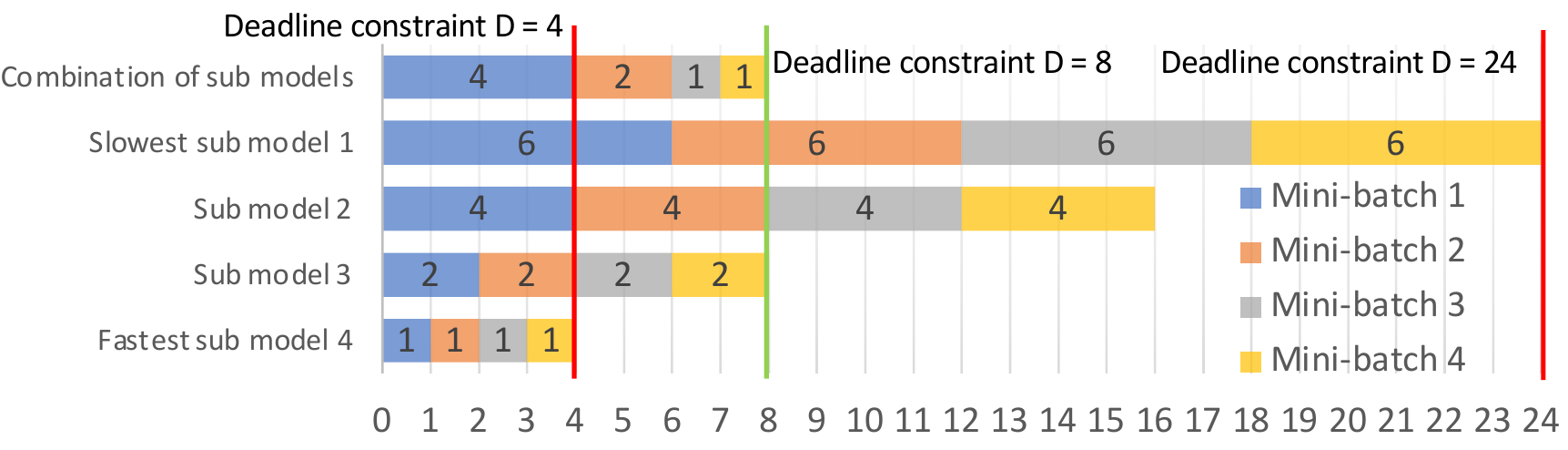}
% \caption{Time spent on processing 4 mini-batches of different sub-models.}
\caption{The total Time required to process all four mini-batches using different sub-models.}
\label{fig:inferencetime}
\end{figure}

In the first scenario where only one single model can be deployed in the system, Algorithm~\ref{algo:sa} is adopted to dynamically adapt the model size to meet the deadline requirement and to obtain
%% ooibc: when you use guarantee, you need proof
%guarantee 
the best \emph{effective accuracy}.
The situations are summarized as follows:
\begin{itemize}
\item
If $D \leq T_{fast}$ (see Figure \ref{fig:inferencetime} where $D \leq 4$), dropping mini-batch will be unavoidable.
In this case, the scheduler will schedule all mini-batches to the fastest sub-model to minimize the drop rate.
% sf: check consistency 'drop rate' <- 'dropping rate'
\item
If $D \geq T_{slow}$ (see Figure \ref{fig:inferencetime} where $D \geq 24$), the scheduler will schedule all mini-batches to the slowest but most accurate sub-model.
\item
If $T_{fast} < D < T_{slow}$ (see Figure \ref{fig:inferencetime} where $4<D<24$), more than one sub-model is needed to achieve the best \emph{effective accuracy}.
\end{itemize}

For a single sub-model $m_i$, it's \emph{effective accuracy} is defined as:

\newcommand{\FS}[2]{\displaystyle\frac{#1}{#2}}
\begin{equation} 
p_i^{eff}=\left\{
\begin{aligned} 
     p_i, && {W_{exp} \leq \widehat{W_i}}  \\ 
     (\widehat{W_i}~/~W_{exp}) * p_i, && {W_{exp} > \widehat{W_i}}\\
\end{aligned} 
\right.
\label{f2}
\end{equation}

When the expected workload $ W_{exp} $ is higher than the maximal workload that a single sub-model can handle, i.e., $ \widehat{W_i} $, the \emph{effective accuracy} decreases since the sub-model cannot process all instances before the deadline.
In this case, the \emph{effective accuracy} can be improved using multiple sub-models. The optimization objective now becomes a combinatorial optimization problem that maximizes the follows:

\begin{equation} 
p^{eff}([n_1,\ldots,n_K]) = \sum_{i=1}^{K}(\FS{n_i}{N_{mb}} * p_i)
\label{fobj}
\end{equation}

Moreover, $n_i$ is the number of mini-batches assigned to the $i$-th sub-model, and at most $ N_{mb} $ mini-batches can be scheduled, we thus have the following bound functions:

\begin{equation} 
n_i \geq 0~(i = 1,2...K)
\label{bd}
\end{equation}
\begin{equation} 
n_i \in \mathbb{Z}~(i = 1,2...K)
\label{bdz}
\end{equation}
where $ \mathbb{Z} $ is the set of integers. We also have:
\begin{equation} 
\sum_{i=1}^{K}n_i \leq N_{mb}
\label{bd2}
\end{equation}
Additionally, the total time to process all of the scheduled mini-batches is limited by $D$:
\begin{equation} 
\sum_{i=1}^{K}(n_i*t_i) \leq D  
\label{tc}
\end{equation}

The maximization of the objective function Eq. \eqref{fobj} with the constraints of Eq. \eqref{bd}, \eqref{bdz}, \eqref{bdz}, \eqref{bd2} and \eqref{tc} can be formulated as a Integer Linear Programming (ILP) problem, which can be solved by either the classical linear programming-based \emph{Branch-and-Bound (B\&B)} method or Dynamic Programming (DP).
Although DP can find the optimal solution in polynomial time, the solution may not be precise as $D$ needs to be discretized, which forms a limited number of sub-problems.
Thus, the B\&B method is used prior to DP, which is shown in Algorithm~\ref{algo:sa}.
Specifically, the B\&B method takes the following steps: 
<s1> An initial linear programming problem $ X_0 $ is constructed by grouping Eq. \eqref{fobj}, \eqref{bd2}, \eqref{bd} and \eqref{tc}. Then $ X_0 $ is pushed to the problem queue $ \phi $. 
<s2> Retrieve a linear programming problem $ X $ from $ \phi $ and get its optimal solution $ \beta $ by applying linear programming method. 
<s3> If all elements in $ \beta $ are integers, then $ \beta $ satisfies Eq. \eqref{bdz} and will be a feasible solution of the ILP problem. If $ \beta $ leads to a higher $ p^{eff} $, then $ \beta $ will be used to update $ \eta $. 
<s4> If $ \beta $ contains non-integers, eg. $ \beta[i] $ is a float number, then two new sub-problems are generated by merging the problem $ X $ and two respective new constraints, namely $ n_i \geq int(\beta[i]) + 1 $ and $ n_i \leq int(\beta[i]) $, which are then pushed into $ \phi $.
Steps <s2>, <s3> and <s4> are iterated until $ \phi $ is empty. Finally, the optimal $ \eta $ can be obtained.
If the B\&B method fails to find the optimal $ \eta $, the ILP problem will be approximated as a classical 2-dimensional unbounded knapsack problem and solved by DP.

Our experiments further confirm that the optimal solution of Eq. \eqref{fobj} is also the best scheduling policy that achieves the highest \emph{effective accuracy}.
For example, as shown in Figure \ref{fig:inferencetime}, when the user sets $D=8$, sub-model 1 can only serve one mini-batch within the deadline.
Sub-models 3 and 4 can meet the deadline but with relatively low accuracy.
The best scheduling policy is a combination of sub-models 2, 3 and 4 that will provide relatively high accuracy and achieve a zero-drop rate.
Specifically, mini-batch 1 is assigned to sub-model 2.
The model is then switched to sub-model 3 to process mini-batch 2 after processing mini-batch 1.
Finally, mini-batch 3 and 4 are processed using sub-model 4.

To further save the time spent on running the scheduling algorithm, we precompute and store combinations of $D$ and $N_{mb}$ in \emph{In-Memory Cache} to accelerate decision making.
%%% ooibc: many? how many
%%% in scientific writing, it is important to be specific

\begin{algorithm}
    \caption{Scheduling Algorithm}
    \label{algo:sa}
    \begin{algorithmic}[1] 
        \Require $\rho = [p_1,\ldots,p_K]$, $\tau = [t_1,\ldots,t_K]$, $D$, $N_{mb}$
        \Ensure $\eta=[n_1,\ldots,n_K]$, theoretical $p^{eff}(\eta)$
        \Function{Scheduler}{$\rho, \tau, D, N_{mb}$}
        \State $\eta \leftarrow [0,\ldots,0] $
        \If {$ D \leq N_{mb} \times \texttt{Min}(\tau) $ }
            $ \eta[\texttt{Argmin}(\tau)] \leftarrow N_{mb} $ 
        \EndIf
        \If {$D \geq N_{mb} \times \texttt{Max}(\tau) $}
            $ \eta[\texttt{Argmax}(\tau)] \leftarrow N_{mb} $ 
        \EndIf
        \If {$ N_{mb} \times \texttt{Min}(\tau) < D < N_{mb} \times \texttt{Max}(\tau) $}
            \State $ \phi \leftarrow \texttt{Queue}([\{{\rm Eq. \eqref{fobj}, \eqref{bd}, \eqref{bd2}, \eqref{tc}}\}]) $
            \While{$ \phi \not = [] $}
                \State $ X \leftarrow \phi.\texttt{Pop} $
                \State $ \beta \leftarrow \texttt{Linear\_Programming}(X) $
                \If {$ p^{eff}(\beta) > p^{eff}(\eta) $}
                    \If {$ \forall i.~ \beta[i] \in \mathbb{Z} $}
                    $ \eta \leftarrow \beta $
                    \Else
                        ~select $ i $ such that $ \beta[i] \not \in \mathbb{Z} $
                        \State $ \phi.\texttt{Push}(X~\cup~\{n_i \geq \texttt{Int}(\beta[i]) + 1\}) $
                        \State $ \phi.\texttt{Push}(X~\cup~\{n_i \leq \texttt{Int}(\beta[i])\}) $
                    \EndIf
                \EndIf
            \EndWhile
            \State //{ If B\&B fails to find a solution} 
            \If {$\eta = [0,\ldots,0]$} 
                \State $ \eta \leftarrow \texttt{2D\_Unbounded\_Knapsack\_DP}(\rho, \tau, D, N_{mb}) $
            \EndIf
        \EndIf
        \State \Return{$\eta$, $p^{eff}(\eta)$}
        \EndFunction
    \end{algorithmic}
\end{algorithm}

% \vspace{-0.1cm}
\subsection{Multiple Model Serving}
\label{sec:multimodelinference}

In the second scenario where multiple models can be loaded to the system, as shown in Figure \ref{fig:inference2scen}(b), the \emph{Producer} will partition instances to different queues.
Each queue is served by a dedicated \emph{Inference Worker}.
The models in other \emph{Inference Workers} are replicated from the first \emph{Inference Worker}.
% The models in other \emph{Inference Workers} are repeated from the model in the first \emph{Inference Worker}.
The model in each \emph{Inference Worker} is able to switch between sub-models.
The global \emph{effective accuracy} is defined as the average of \emph{effective accuracy} of each \emph{Inference Worker}.
Suppose there are $b$ \emph{Inference Workers}, the \emph{global effective accuracy}=$\frac{1}{b} \sum_{j=1}^{b}p_j^{eff}$.
The global best \emph{effective accuracy} is equal to the average of the local best \emph{effective accuracy}.
To get the best local \emph{effective accuracy}, each \emph{Inference Worker} runs Algorithm~\ref{algo:sa} separately under the original deadline constraint $D$ and the number of mini-batches $ N_{mb} $ in the corresponding queue.

% sf: update this subsection as discussed

% \vspace{-0.05cm}
\section{System Usability}
\label{sec:userInteraction}

\begin{figure}
\centering
\includegraphics[width=8cm]{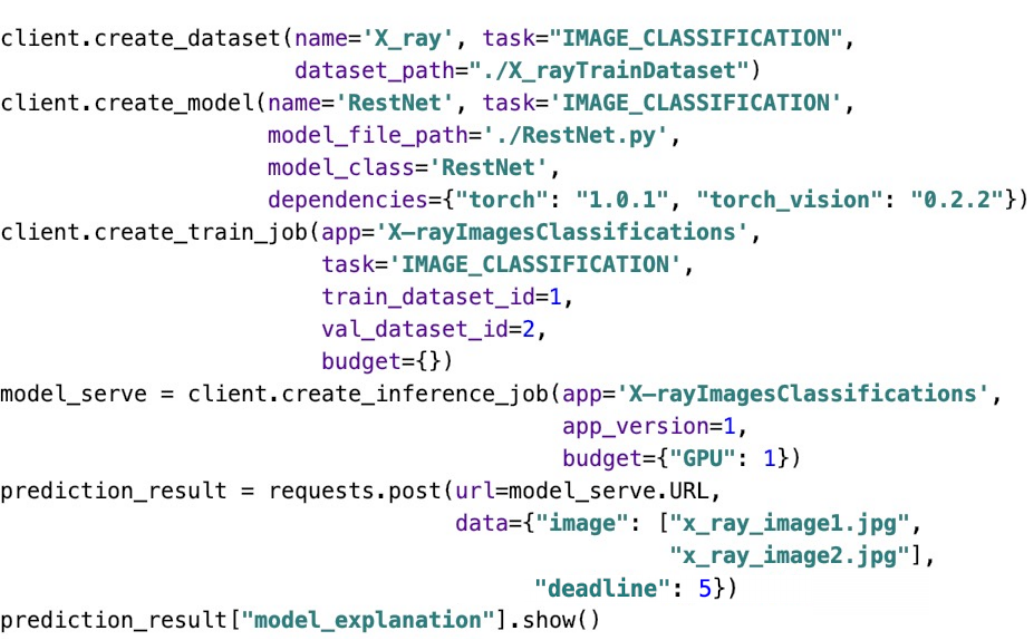}
% \caption{\system{} APIs for creating X-ray image classification.}
\caption{\system{} APIs for supporting X-ray image classification.}
\label{fig:foodcode}
\end{figure}

To improve the usability of ML and DL models in multimedia applications such as medical image classification, food recognition, dietary management, question answering, and speech classification, \system{} provides built-in models using third-party libraries based on PyTorch~\cite{paszke2019pytorch}, TensorFlow~\cite{abadi2016tensorflow} and Scikit-learn~\cite{DBLP:journals/jmlr/PedregosaVGMTGBPWDVPCBPD11}. 
Table \ref{tab:ml-models} lists representative models for six common multimedia tasks.

Figure \ref{fig:foodcode} shows an example on the use of \system{}'s APIs to quickly develop applications with the supported models. To improve the reliability of inference, we have also implemented two model explanation techniques, i.e., Grad-CAM~\cite{selvaraju2017grad} and LIME~\cite{ribeiro2016should}.

%% ooibc; by writing this way, it changes the scope of what is being said
%The experimental results are presented in Section ~\ref{sec:usereval}}

\section{Experimental Evaluation}
\label{sec:experiment}

To evaluate the usability and adaptability of \system{}, we conduct experiments on different multi-media datasets using various DL network architectures. \system{} is deployed on clusters equipped with GTX1080Ti GPUs and two models of CPUs. The CPU model deployed in \emph{Admin} node is Intel Xeon CPU E5-2620 v4 @ 2.10GHz. The CPU model deployed in the other nodes is Intel Xeon CPU E5-1650 v4 @ 3.60GHz.
All nodes are connected via Ethernet at 1 Gbit/s.
We run all services in Docker and use Kubernetes 1.6 as the cluster manager.

% \vspace{-0.5cm}
% \subsection{Training Evaluation}
\subsection{Experimental Setup}
\label{sec:traineval}

In this section, we introduce the training details, namely the datasets adopted for the evaluation and the training results.

\begin{figure}[t!]
% \setlength{\abovecaptionskip}{-0.1cm}
% \setlength{\belowcaptionskip}{-0.5cm}
% \vspace{-0.2cm}
\centering
    \subfigure[Inference time of different sub-models.]{
        \begin{minipage}[a]{0.4\textwidth}
        \centering
        \includegraphics[width=6cm]{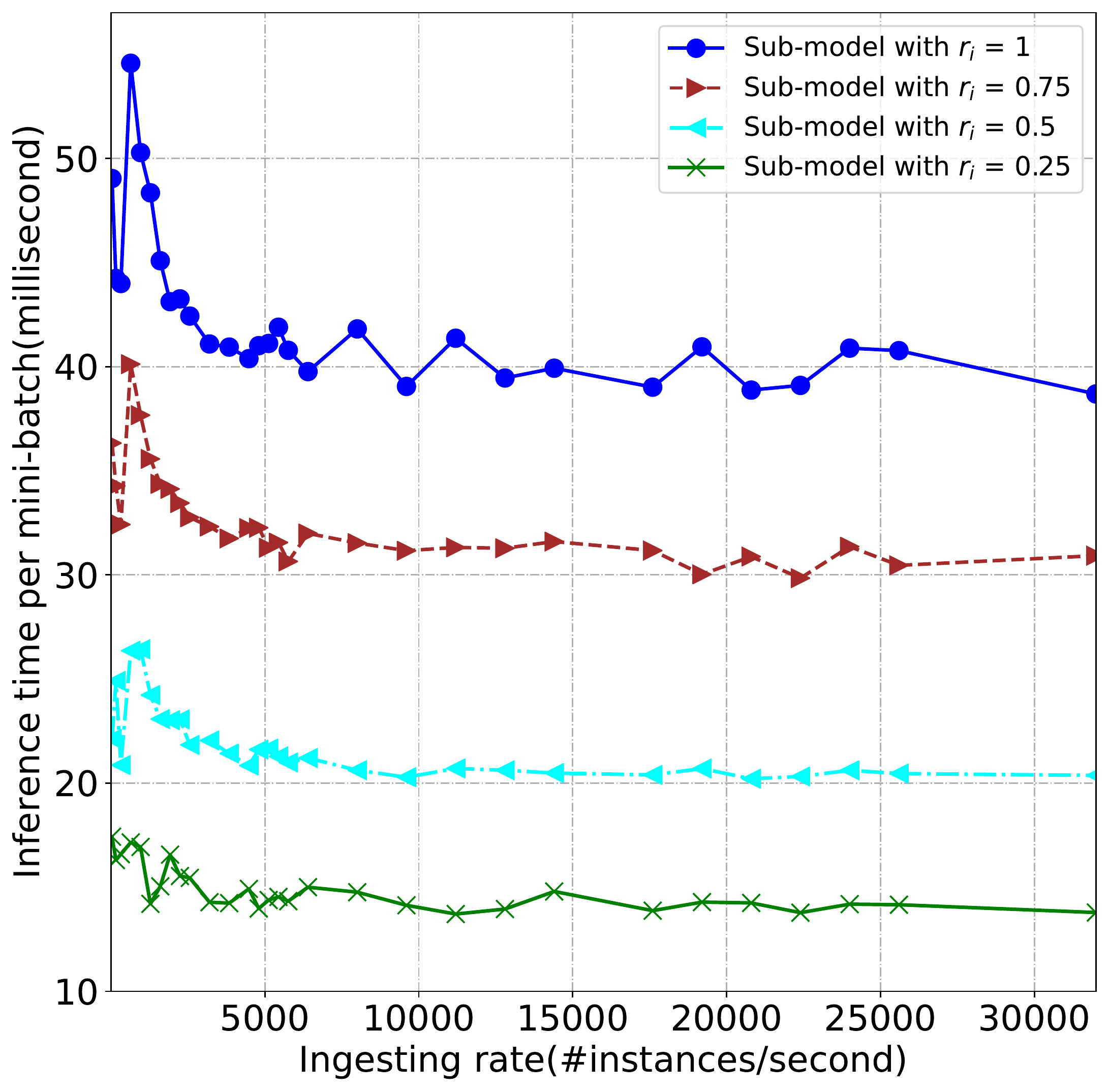}
        \end{minipage}}
    \subfigure[Illustration of the impact of the scheduling algorithm on \emph{Effective Accuracy} (ResNet-50 trained with model slicing on NIH Chest X-rays).]{
        \begin{minipage}[b]{0.4\textwidth}
        \centering
        \includegraphics[width=6cm]{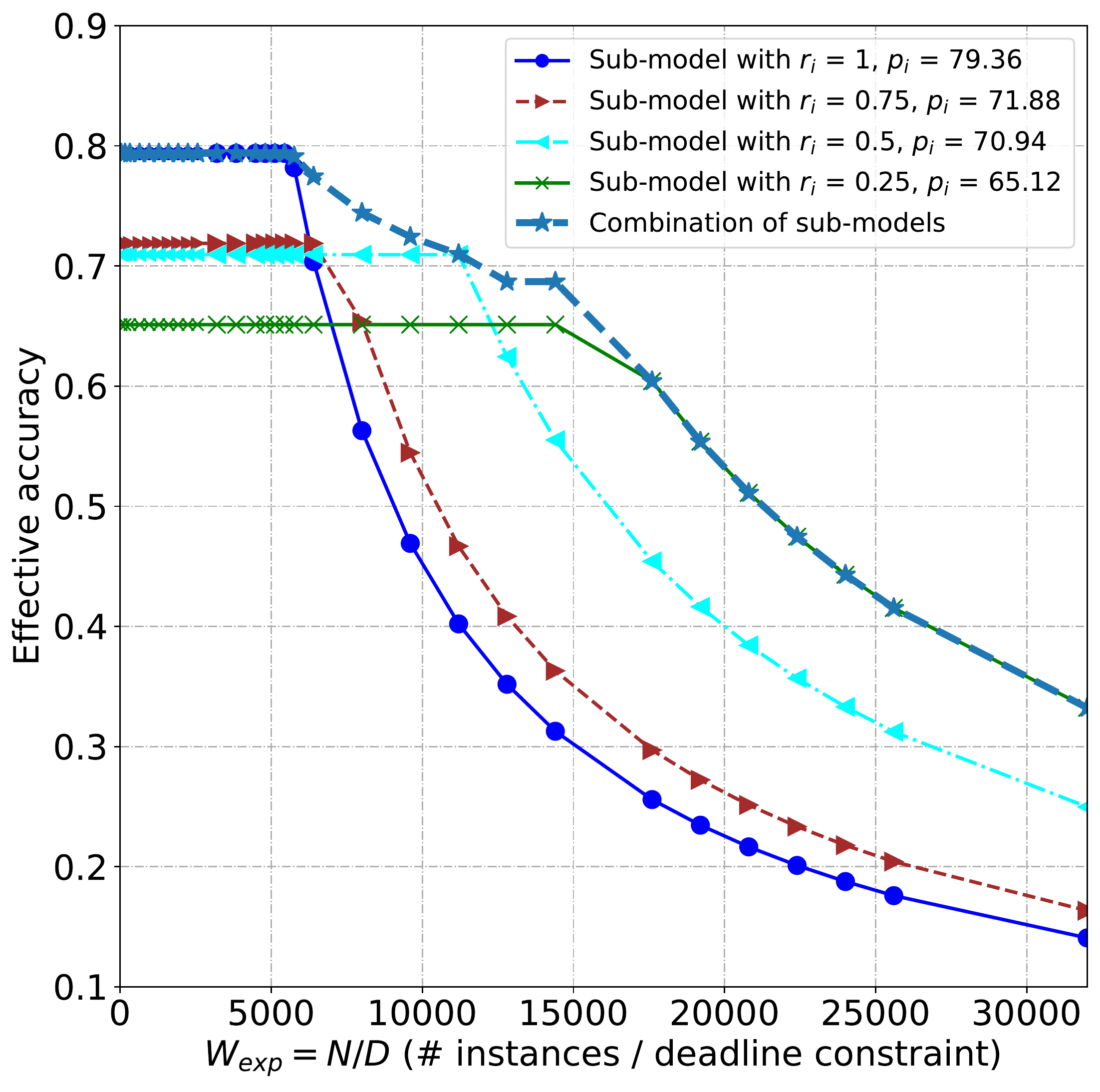}
        \end{minipage}}
\centering
\caption{Inference time and \emph{effective accuracy}.}
\label{fig:effeccomp}
\end{figure}

\begin{figure*}[t!]
% \setlength{\abovecaptionskip}{-0.1cm}
% \setlength{\belowcaptionskip}{-0.5cm}
% \vspace{-0.2cm}
\centering
\subfigure[Throughput]{
    \begin{minipage}[t]{0.30\linewidth}
    \centering
    \includegraphics[width=5.4cm]{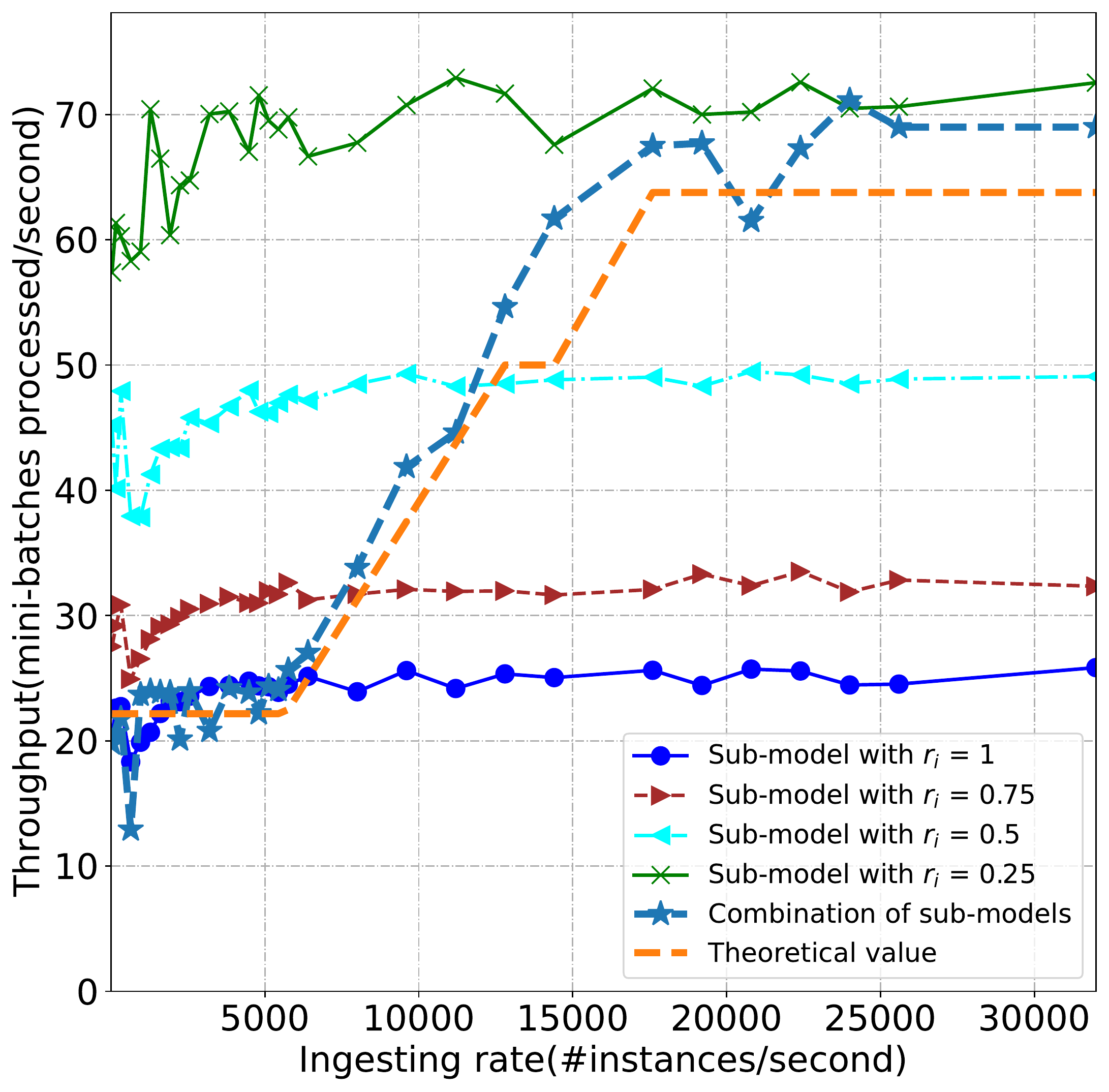}
    \end{minipage}}
\hspace{0.005\textwidth}    
\subfigure[Tail latency]{
    \begin{minipage}[t]{0.30\linewidth}
    \centering
    \includegraphics[width=5.4cm]{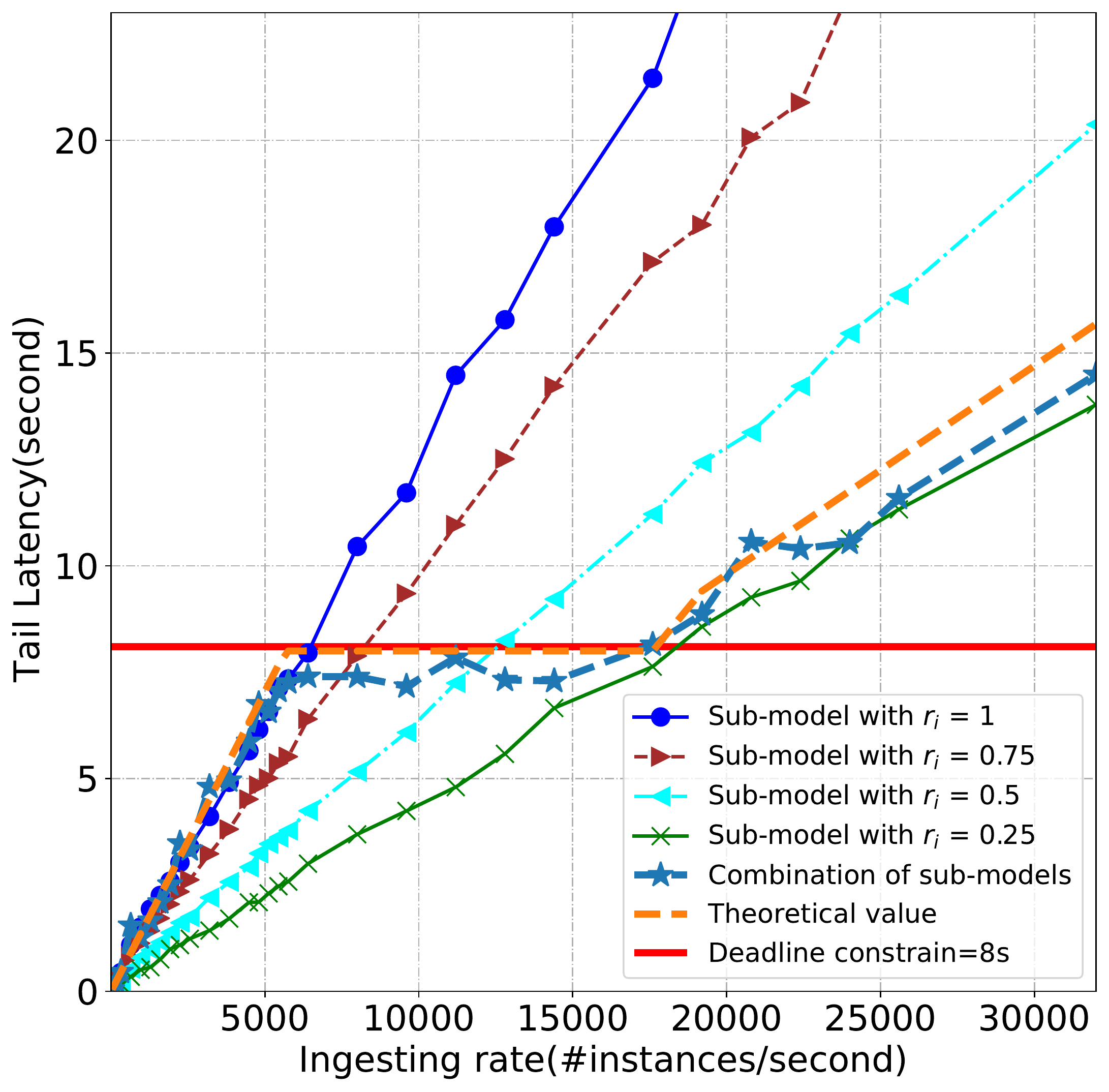}
    \end{minipage}}    
\hspace{0.005\textwidth}    
\subfigure[Ingesting rate on scheduled sub-models.]{
    \begin{minipage}[t]{0.30\linewidth}
    \centering
    \includegraphics[width=5.75cm]{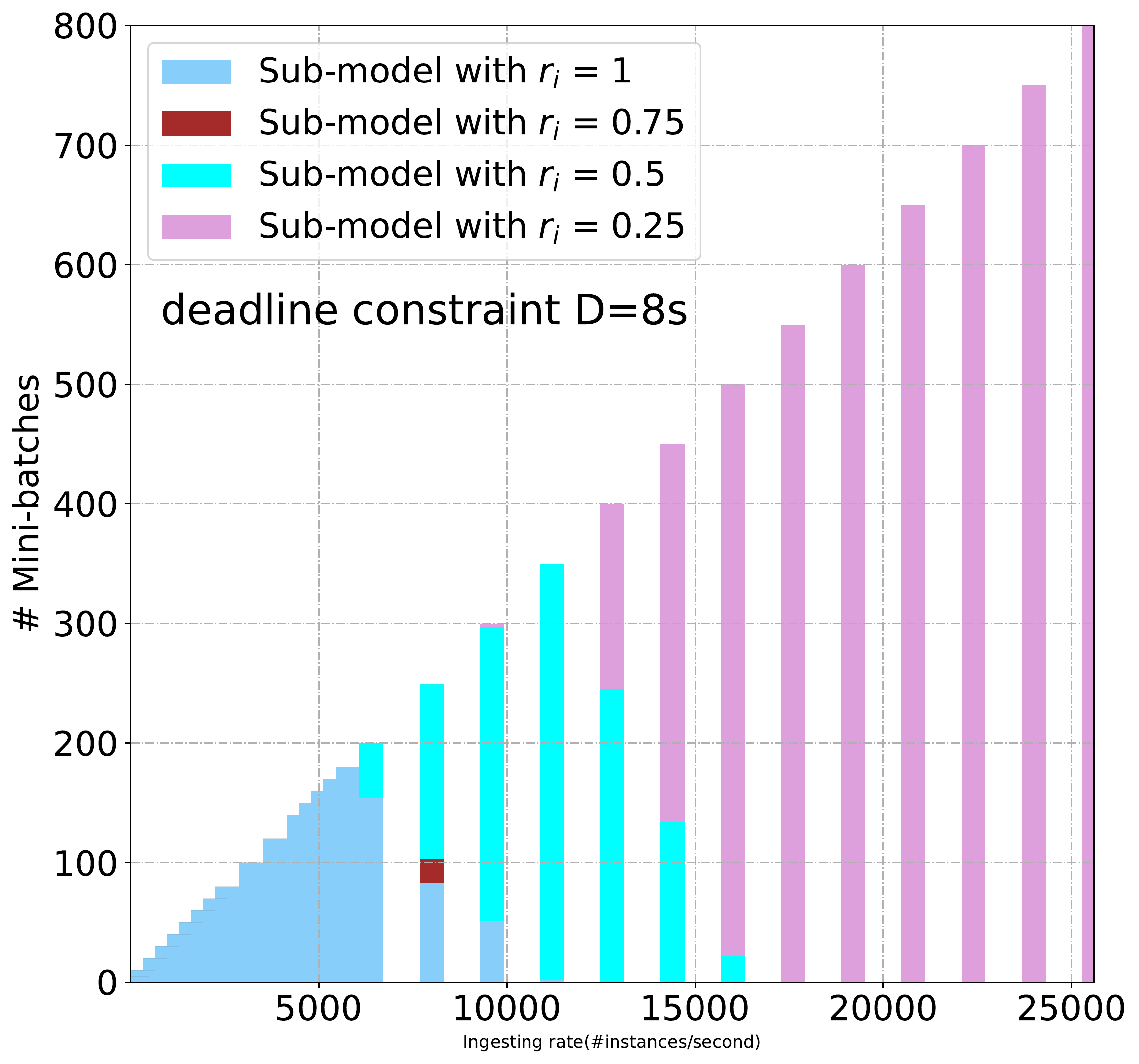}
    \end{minipage}}
\centering
% \caption{Experimental results exhibit similar trend as the theoretical values.}
\caption{Adaptability Experiments in \system{}}
% sf: change the caption
\label{fig:combinesthree}
\end{figure*}

\subsubsection{Datasets}
\label{sec:datasetforinf}

\begin{table}[h!]
  \caption{Models for multimedia applications.}
  \label{tab:ml-models}
  \begin{tabular}{ll}
  \toprule
  Model & Scenario \\
  \midrule
  VGG~\cite{DBLP:journals/corr/SimonyanZ14a}, ResNet~\cite{DBLP:conf/cvpr/HeZRS16} & Image Classification \\
  Mask RCNN~\cite{DBLP:conf/iccv/HeGDG17} & Object Detection \\
  Deep Speech~\cite{10.5555/3045390.3045410} & Speech Recognition \\
  BERT~\cite{DBLP:conf/naacl/DevlinCLT19} & Question and Answering \\
  Random Forest~\cite{DBLP:conf/icdar/Ho95} & Tabular Classification \\
  XGBoost~\cite{DBLP:conf/kdd/ChenG16} & Tabular Regression \\
  \bottomrule
\end{tabular}
% \vspace{-0.3cm}
\end{table}

We evaluate the training and inference efficiency of \system{} using model ResNet-50 on three image classification datasets, namely CIFAR~\cite{cifar10cite}, ILSVRC 2012~\cite{imagenetcite} and NIH Chest X-rays dataset\footnote{NIH Chest X-rays: www.kaggle.com/nih-chest-xrays/data}. 
To demonstrate the general support of \system{} for different applications, we train Inception-ResNet-v2 and yoloV3~\cite{yolov3} on five food datasets for food image classification and detection and visualize the results.
We use 50,000 training images and 10,000 test images from the CIFAR dataset. Each CIFAR image is resized to 32 $ \times $ 32.
We use 1.2 million training images and 50,000 test images drawn from 1,000 classes in the ILSVRC 2012 dataset. And resize them to 244 $ \times $ 244.
We also use 5,234 training images and 634 test images from the NIH Chest X-rays dataset.
Each image of the X-rays dataset is classified as "healthy" or "unhealthy" and is normalized from 1600 $ \times $ 1125 to 244 $ \times $ 244.
We also use five Singapore food datasets.
The number of classes in each dataset is 55, 101, 172, 231 and 256, respectively.
Each class contains 300 to 500 images of size 624 $ \times $ 1024.

\subsubsection{Training Results}
\label{sec:trainResSection}

\begin{figure*}[ht]
% \setlength{\abovecaptionskip}{-0.0cm}
% \setlength{\belowcaptionskip}{-0.3cm}
% \vspace{-0.4cm}
\includegraphics[width=17cm]{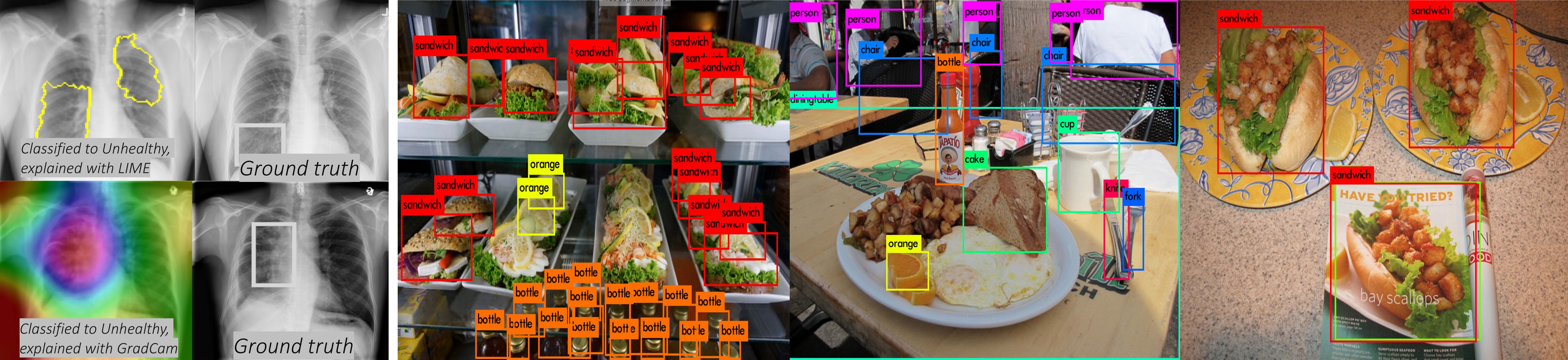}
\caption{Two multimedia applications(X-ray based diagnosis and Food Detection) developed using \system{}. The correctness of X-ray explanation is confirmed by the overlap between the explanation map and the ground truth.} 
\label{fig:multiapps}
\end{figure*}

ResNet-50 is trained on each dataset with SGD.
Specifically, we train 100/100/300 epochs on CIFAR-10/ILSVRC 2012/NIH X-arays with a batch size of 128/64/64, respectively.
We summarize the statistics of the trained models in Table ~\ref{tab:trainres}.

\begin{figure}
\centering
\includegraphics[width=8cm]{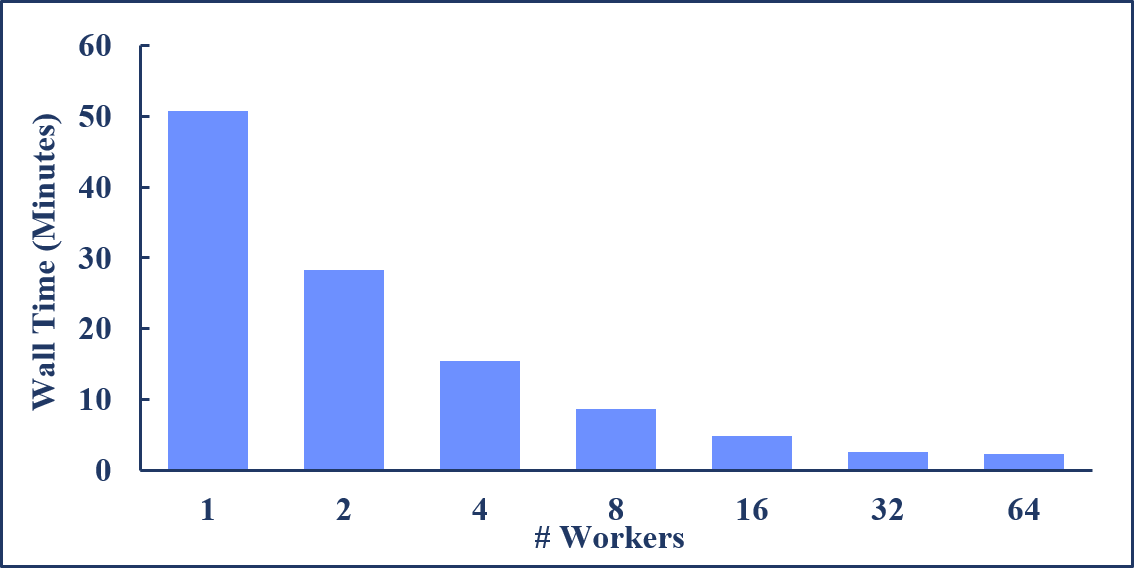}
\caption{Scalability test of multiple workers.}
\label{fig:cluster-test}
\end{figure}

%NL: this is comment due to page limit 6-8 pages
In order to demonstrate the scalability of \system{}, we enabled 2, 4, 8, 16, 32 and 64 parallel workers in a cluster and tested the wall time for the parallel workers to complete 64 training jobs in total. Each worker was allowed to use at most 12 logical processors and 32GB of memory. For each training job, a multi-layer perceptron with random hyper-parameters, i.e. the initial learning rate, the number of hidden layers and the dimensionality of hidden layers, was trained using 1,000 gray images drawn from 10 classes. Each image was of size $ 50 \times 50 $ pixels. Figure \ref{fig:cluster-test} shows the wall time with respect to the number of parallel workers. It demonstrated that multiple training jobs can be effectively run in parallel in our system.

% \vspace{-0.1cm}
\subsection{Dynamic Model Serving Evaluation}
\label{sec:dynamic_eval}

The adaptability of \system{} is evaluated using ResNet-50 trained on dataset NIH Chest X-rays.
In the following experiments, we set the mini-batch size $S_{mb}$ to $ 32 $ for the evaluation.

We first measure the inference time $ t_i $ to process a single mini-batch with different sub-models.
Then we measure the \emph{effective accuracy} of \system{} under the first scenario where only one single model can be loaded to the system.
%%%ww: -->where there is only a single model?
%%NL: this is the situation where the resource is only allowed to load a single model

To measure the actual inference time, we enable GPU warm-up and GPU/CPU synchronization.
%%%ww: what's the synchronization?
%% NL: it basically call torch.cuda.synchronize()..
We also use \emph{torch.cuda.Event} to capture the time before and after model inference.
Specifically, we record the inference time with different ingesting rates from 32 to 25,000 instances/second as shown in Figure ~\ref{fig:effeccomp}(a).
Then the inference time is averaged to obtain the $ t_i $ of sub-model $ m_i $.
Results in Table ~\ref{tab:trainres} show that both the accuracy and inference time decreases with a smaller slice rate, which is consistent with the previous discussion.

\begin{table}[t!]
\label{T:equipos}
\caption{Accuracy and inference time of sliced ResNet-50.}
\label{tab:trainres}
        \begin{tabular}{|p{0.8cm}<{\centering}|p{0.8cm}<{\centering}|p{0.8cm}<{\centering}|p{0.8cm}<{\centering}|p{0.8cm}<{\centering}|p{0.8cm}<{\centering}|p{0.8cm}<{\centering}|}
        \hline
        \textbf{Slice}
        & \multicolumn{2}{c|}{\textbf{CIFAR-10}} & \multicolumn{2}{c|}{\textbf{ImageNet12}} & \multicolumn{2}{c|}{\textbf{X-Ray}} \\
        \cline{2-7}
        \textbf{Rate}
        & \textbf{Acc.~} & \textbf{$T_{ave}$} & \textbf{Acc.~} & \textbf{$T_{ave}$} & \textbf{Acc.~} & \textbf{$T_{ave}$} \\
        \hline
        1    & 91.13 & 12.48 & 75.09 & 49.40 & 79.37 & 45.12\\ \hline
        0.75 & 88.41 & 9.92 & 73.74 & 38.08 & 71.88 & 34.56 \\ \hline
        0.5  & 85.19 & 6.41 & 71.09 & 22.95 & 70.94 & 22.72 \\ \hline
        0.25 & 79.71 & 3.24 & 63.91 & 17.82 & 65.12 & 15.68\\ \hline
        \multicolumn{7}{l}{*$T_{ave}$: average inference time (ms) to process a mini-batch.}
        \end{tabular}
% \vspace{-0.5cm}        
\end{table}

To measure the \emph{effective accuracy} of \system{}, we set the deadline constraint to $ D=8s $ and gradually increase the number of ingested instances $N$ from 32 to 30,000.
As shown in Figure ~\ref{fig:effeccomp}(b), the model equipped with the scheduler can adapt to the workload by switching between sub-models, which leads to higher \emph{effective accuracy}. Specifically, when $ W_{exp}=100 $, the serving model is the full model (i.e., the slice rate $r=1.0$).
When $ W_{exp}=10,000 $, the serving model is switched to the smaller model of a slice rate 0.5 to avoid dropping instances. When $ W_{exp}=18,000 $, where even the smallest sub-model of a slice rate 0.25 can not process all instances within the time limit.
In such scenarios, the serving model is switched to the smallest sub-model to maximize the throughput.
As shown in Figure ~\ref{fig:combinesthree}(a), the system dynamically adapt the model size to increase the throughput until it reaches the maximum throughput, which is the same as the fastest sub-model.

We also measure the latency of the system, which is shown in Figure ~\ref{fig:combinesthree}(b).
Specifically, when the ingesting rate is low, since all sub-models now can meet the deadline, the scheduler will adopt the sub-model of slice-rate 1.0 for higher accuracy.
When the ingesting rate reach around 10,000, both the sub-models of a slice-rate 1.0 and 0.75 cannot process all instances before the deadline.
The combination of sub-models however, can meet the deadline constraint until the ingesting rate reach 18,000, where the serving model will entirely switch to the sub-model of a slice rate 0.25.

To better illustrates the combinations of the scheduled sub-models under different instance ingesting rates, we present detailed assignment of the mini-batches to the sub-models in Figure ~\ref{fig:combinesthree}(c).
We can observer that when the ingesting rate is low, the model assigns all mini-batches to the sub-model of a slice rate 1.0.
Since the sub-model of a slice rate 0.75 and 0.5 have similar accuracy, while the sub-model of a slice rate 0.5 is much faster, the scheduling algorithm tends to use the later sub-model for achieving higher \textit{effective accuracy}.

For the second scenario, where multiple models can be loaded to the system, \system{} can have multiple elastic models and can generate multiple combinations of sub-models. In contrast,  Model-Switching can only have fixed combinations of models. 
% sf: update this part accordingly

In conclusion, the experiments on \emph{effective accuracy}, throughput, latency, and sub-model combinations confirm that the model trained with the model slicing technique and our proposed scheduling algorithm support dynamic workloads via finer-grained elastic computation control. 
It further illustrates the adaptability of \system{}.

%%% ooibc: use a better caption!!!
%% ooibc: why are the food images (centre one expecially) so complicated?
%%% nl: the food image is an object detection task, each box is bounding box detected.
%%% nl: the x-ray image is a classification task with explanation display.

\subsection{Multimedia Applications}
\label{sec:usereval}

We further demonstrate the usability of \system{} on various applications.
Due to the space limit, we showcase representative examples in Figure ~\ref{fig:multiapps}.
The Singapore Food Detection component has been used to develop FoodLG app\footnote{http://foodlg.com/}, which is customized for healthcare applications such as pre-diabetes management and diet recommendation.
For the training dataset, we crowdsource to knowledge users using CDAS\cite{cdas} for labelling.
Medical applications like X-ray-based diagnosis is shown in Figure ~\ref{fig:multiapps}, the GradCam map highlights the unhealthy areas with warm colors (red and purple).
The LIME map circles the unhealthy areas with yellow color.
The explanation maps can assist clinicians in verifying the correctness of the diagnosis, e.g., whether explanation maps match is in line with their diagnosis.

% \vspace{-0.01cm}
\section{Related Work}
\label{sec:related}

In this section, we review the related work of ML/DL systems and framework. Their are highly accessible and could be used to extend our \system{}.

PyTorch~\cite{paszke2019pytorch} can achieve automated ML using the Auto PyTorch library~\cite{DBLP:journals/corr/abs-2006-13799}, but it does not provide the system infrastructure for ML life cycle management in multimedia applications. \system{} can be used to facilitate the PyTorch models.

Microsoft NNI\footnote{Microsoft NNI: https://github.com/microsoft/nni} is a ML framework supporting model compression. However, it does not provide elastic inference capabilities to the models. While the slice-rate in \system{} is more understandable.

Hopswork~\cite{ismail2017hopsworks} is a data science platform for the design and operation of data analytics applications. The system applies HopsFS, a highly scalable distributed file system, to improve its efficiency. while our system focuses more on the usability to AI applications.

In summary, there are indeed many data analytics systems developed in recent years. Our \system{} is designed to improve the usability and adaptability in developing multimedia applications.

% \vspace{-0.2cm}
\section{Conclusions}
\label{sec:conclusions}

In this paper, we introduced \system{} - a learning system focusing on usability and adaptability. \system{} was built on top of Apache SINGA. It assists users in managing data and models, and developing AI applications. We have used \system{} to develop multi-media applications such as a chest X-ray image explanation function and food detection system. 
We showed that \system{} is highly extendable as it can be used with various third-party machine learning models.

Moving forward, we note that there exist other bottlenecks in data science such as data loading, visualization, cleaning, labeling, and data transformation.
Future extensions to \system{} may include such data science supporting modules.

% \vspace{2mm}
\textbf{Acknowledgement:} We thank the anonymous reviewers for their constructive comments and 
% Shaofeng Cai for his comments and fundamental contributions.
NUS colleagues for their comments and contributions.
This research is supported by Singapore Ministry of Education Academic Research Fund Tier 3 under MOE's official grant number MOE2017-T3-1-007.  Meihui Zhang's work is supported by the National Natural Science Foundation of China (62050099).

\bibliographystyle{ACM-Reference-Format}
\bibliography{ref-clean}

\end{document}